\documentclass[sigconf]{acmart}
\AtBeginDocument{%
  \providecommand\BibTeX{{%
    \normalfont B\kern-0.5em{\scshape i\kern-0.25em b}\kern-0.8em\TeX}}}

\setcopyright{none}
\copyrightyear{2021}
\acmYear{2021}

\acmConference[HRI]{HRI '21: ACM Int. Conf. Human Robot Interaction}{Marh 8--11, 2021}{Boulder, CO}

\copyrightyear{2021}
\acmYear{2021}
\setcopyright{acmcopyright}\acmConference[HRI '21]{Proceedings of the 2021 ACM/IEEE International Conference on Human-Robot Interaction}{March 8--11, 2021}{Boulder, CO, USA}
\acmBooktitle{Proceedings of the 2021 ACM/IEEE International Conference on Human-Robot Interaction (HRI '21), March 8--11, 2021, Boulder, CO, USA}
\acmPrice{15.00}
\acmDOI{10.1145/3434073.3444657}
\acmISBN{978-1-4503-8289-2/21/03}

\usepackage{caption}
\usepackage{subcaption}
\usepackage{amsfonts}
\usepackage{multirow}
\usepackage{color}
\usepackage[normalem]{ulem}
\usepackage{graphicx}

\graphicspath{ {../} }

\newenvironment{flushitemize}{%
\begin{list}{$\bullet$}
   {\setlength{\leftmargin}{15pt}}%
    \setlength{\labelwidth}{20pt}
    \setlength{\itemindent}{0pt}
    \setlength{\labelsep}{0.5em}
 \setlength{\itemsep}{1pt}
 \setlength{\parskip}{0pt}
 \setlength{\parsep}{0pt}}
 {\end{list}}

\begin{document}

\title{Explainable AI for Robot Failures: Generating Explanations that Improve User Assistance in Fault Recovery}

\author{Devleena Das}
\affiliation{%
 \institution{Georgia Institute of Technology}
 \country{Atlanta, Georgia}}
 \email{ddas41@gatech.edu}
 
 \author{Siddhartha Banerjee}
\affiliation{%
 \institution{Georgia Institute of Technology}
 \country{Atlanta, Georgia}}
 \email{siddhartha.banerjee@gatech.edu}
 
\author{Sonia Chernova}
\affiliation{%
 \institution{Georgia Institute of Technology}
 \country{Atlanta, Georgia}}
 \email{chernova@gatech.edu}



\begin{abstract}
With the growing capabilities of intelligent systems, the integration of robots in our everyday life is increasing. However, when interacting in such complex human environments, the occasional failure of robotic systems is inevitable. The field of explainable AI has sought to 
make complex-decision making systems more interpretable but most existing techniques target domain experts. On the contrary, in many failure cases, robots will require recovery assistance from \textit{non-expert} users. In this work, we introduce a new type of explanation, $\mathcal{E}_{err}$, that explains the cause of an unexpected failure during an agent's plan execution to \textit{non-experts}. In order for $\mathcal{E}_{err}$ to be meaningful, we investigate what types of information within a set of hand-scripted explanations are most helpful to non-experts for failure and solution identification. Additionally, we investigate how such explanations can be autonomously generated, extending an existing encoder-decoder model, and generalized across environments. We investigate such questions in the context of a robot performing a pick-and-place manipulation task in the home environment. Our results show that explanations capturing the \textit{context} of a failure and  \textit{history} of past actions, are the most effective for failure and solution identification among non-experts. Furthermore, through a second user evaluation, we verify that our model-generated explanations can generalize to an unseen office environment, and are just as effective as the hand-scripted explanations.


\end{abstract}



\maketitle

\fancyhead{} 
\section{Introduction}

In homes, hospitals, and manufacturing plants, robots are increasingly being tested for deployment alongside non-roboticists to perform complex tasks, such as folding laundry \cite{yang2016repeatable}, delivering laboratory specimens \cite{bloss2011mobile,hu2011advanced}, and moving inventory goods \cite{hagele2016industrial,lawton2016collaborative}. 
When operating in such complex human environments, occasional robot failures are inevitable. When failures occur, human assistance is often required to correct the problem \cite{bauer2008human}, and co-located users -- homeowners, medical lab technicians, and warehouse workers -- will be first on the scene. We classify such users as \textit{everyday users}, or \textit{non-experts}, because of their lack of formal training in machine learning, AI, or robotics.

In order for everyday users to be able to assist in robot failure recovery, they will need to understand \textit{why} a failure has occurred. For example, a homeowner waiting for a robot to bring them coffee may need to determine why the robot suddenly stopped in the middle of the kitchen, or a production line worker may need to determine why a packing robot suddenly stopped picking up items.


The field of Explainable AI (XAI) has sought to address the challenge of understanding "black-box" systems through the development of interpretable machine learning (ML) algorithms that can explain their decision making to users \cite{gunning2019darpa,adadi2018peeking}.  Furthermore, a subfield of XAI, Explainable Planning (XAIP), has focused on generating explanations specifically for sequential-decision making tasks, including explaining an agent's chosen plan and explaining unsolvable plans to end-users \cite{chakraborti2020emerging}. Such techniques hold great promise for the development of more transparent robotic systems but they do not incorporate explanations for unexpected failures \textit{during} a plan execution. Additionally, the majority of existing XAI techniques are designed for technical domain experts who understand AI or ML at its core \cite{selvaraju2017grad, wu2017beyond,adadi2018peeking,zhang2019interpreting, ribeiro2016should}.  While expert understanding is crucial, such XAI methods are not suitable for the vast majority of end-users, who are non-experts \cite{ehsan2019automated, kambhampati2019synthesizing, das2020leveraging}.

In this work, we seek to make robotic systems more transparent to their users by leveraging techniques from explainable AI, while also extending the capabilities of XAI systems toward greater transparency for non-expert users. Specifically, our work addresses fault recovery cases in which the robot's task execution is halted due to an error. We investigate whether providing explanations can not only help non-expert users understand the system's point of failure, but also help them determine an appropriate solution required to resume normal operation of the task.In some cases (e.g., complex hardware failures), the user might not have the knowledge to fix the point of failure regardless of the provided error explanation. In this work, we focus on failures that we expect to be within the user’s understanding (e.g., object is too far away), and we address this question in the context of pick-and-place manipulation tasks in the home environment. Our work makes the following contributions:
\begin{flushitemize}
    \item \textbf{Formalization of error explanations:} Providing justifications for points of failures that occur unexpectedly \textit{amidst} an agent's plan execution has not previously been studied within the XAIP community. We expand upon the existing set of explanations available in the XAI and XAIP community, introducing \textit{error explanations} designed to explain failures that occur during the execution of a task.
    
    \item \textbf{Explanation content:} We empirically evaluate what information an error explanation should contain to aid non-experts in understanding the cause of failure and to select a recovery strategy. We show that explanations that include both (i) history of recently accomplished actions, and (ii) contextual reasoning about the environment, are the most effective in enabling users to identify the cause of and solution to a failure.
    
    \item \textbf{Explanation generation:} We present an automated technique for generating natural language error explanations that rationalize encountered failures in a manner that is understandable by non-experts. Specifically, we extend an encoder-decoder model for autonomously generating natural language explanations introduced by \cite{ehsan2019automated} to generate context-based history explanations within a continuous state-space.
    
    \item \textbf{Validation with non-expert users:} We demonstrate that explanations generated by the encoder-decoder model can generalize to an unseen environment and are as effective as hand-scripted, context-based history explanations.
\end{flushitemize}

We validate our approach through two user studies and computational model analysis. In the first study, we examine what information an error explanation should contain by evaluating how the content of an explanation affects user performance in identifying and assisting with a robot error (Sec.~\ref{sec:information-properties}). From these results, we identify an explanation type that leads to the highest performance, and then contribute a computational model to automatically generate such explanations from robot states (Sec.~\ref{sec:automated-explanations}). In our second study, we demonstrate that our automated explanations are as effective as hand-scripted explanations in guiding non-experts to identify the cause of a failure and its potential solution.

\section{Related Works}
\label{sec:related-work}

The XAI community has primarily focused on developing interpretability methodologies for understanding the inner workings of black-box models \cite{adadi2018peeking,ribeiro2016should}. Many of these approaches have  focused on model-agnostic implementations, designed to increase \textit{expert} understanding of deep learning outputs \cite{adadi2018peeking,raman2012explaining,rai2020explainable}. Additionally, most XAI approaches  \cite{selvaraju2017grad,zhang2019interpreting,wu2017beyond,ribeiro2016should} have primarily focused on understanding classification-based tasks. However, classification tasks do not capture the complexity of sequential decision-making an agent, such as a robot, may perform while having long-term interactions with users \cite{chakraborti2020emerging}.



To address the need for interpretable explanations in sequential decision making tasks, the XAIP community has focused on explaining an agent's plans to end-users. A recent survey paper highlights some of the key components of plan explanations studied by the community~\cite{chakraborti2020emerging}: (1) contrastive question-answering, (2) explaining unsolvable plans, and (3) providing explicable justifications for a chosen plan. In the realm of contrastive question-answering, Krarup et al. provide a framework to transfer domain-independent user questions into constraints that can be added to a planning model~\cite{krarup2019model}, while Hoffmann et al. utilize common properties within a set of correct plans as an explanation for unmet properties in incorrect plans \cite{hoffmann2019explainable}. In order to explain unsolvable plans, Sreedharan et al. abstract the unsolvable plan into a simpler example through which explanations can be formulated \cite{sreedharan2019can}. Finally, in order to provide explicable justifications for a plan, Zhang et al. use conditional random fields (CRFs) to model human explanations of existing agent plans, and use such a human ``mental model'' as a constraint for generating explicable plans~\cite{zhang2017plan}. The work is extended by Chakraborti et al., who eschew constraining an agent's plan and instead achieve explicability through model reconciliation, whereby the agent provides explanations that reconcile its model to the human ``mental model'' ~\cite{chakraborti2019explicability,chakraborti2017plan}. However, in these works, an explanation justifies a chosen plan, or the lack of one. In our work, we aim to explain the possible failures that can arise \textit{during} a plan.



Techniques for \textit{plan repair} enable a task plan to be adapted to overcome an error~\cite{hammond1990explaining,chang1993comparison}. Methods in this domain reuse the failing plan and search the plan space to find local deviations that allow continued execution~\cite{chen2020rprs}, or transform the plan to adapt it to the situation~\cite{kazhoyan2020towards,nair2020feature}. Such repairs are often found and executed autonomously, with no human intervention.

Recently, works have considered interactive plan repair with a human-in-the-loop. Boteanu et al. showed a proof-of-concept model in which a human approved repair action is proposed by a common-sense reasoning framework~\cite{boteanu2015towards}. However, this work was limited to errors involving missing items, and did not focus on explaining errors to users. Meanwhile, Knepper et al. investigated the grounding of natural language requests to best garner help from non-expert humans~\cite{knepper2015recovering}. They found that the requests were successful when they were targeted, e.g. helped listeners disambiguate between multiple objects, and told them what to do. The authors developed a system to generate such requests. We build upon these findings to investigate the characteristics of natural language error explanations that allow non-experts to help a robot; unlike~\cite{knepper2015recovering}, we do not assume the robot is aware of a correct recovery and able to direct the user on the recovery process.

Plan repair for failures that occur during execution require a fault diagnosis~\cite{hammond1990explaining} and there is a large body of ongoing work in robotics focused on fault diagnosis techniques~\cite{khalastchi2018fault}. These works use a range of methods, including unsatisfied preconditions~\cite{chen2020rprs,boteanu2015towards,knepper2015recovering}, first-order logic inference~\cite{Zaman2013}, case-based reasoning~\cite{Parker2006,hammond1990explaining}, sensor signal processing~\cite{Crestani2015,khalastchi2018diag,abci2020informational}, Bayes nets and Dynamic Bayesian Networks~\cite{Kirchner2014,beck2015skill}, Hidden Markov Models (HMMs)~\cite{Wu2018a}, particle filters~\cite{Verma2004,zhang2019fault}, and neural networks~\cite{Pettersson2007,Park2017} to diagnose failures. Depending on the context of the work, the diagnosis either identifies \textit{what} is wrong with the robot---e.g., object not visible~\cite{knepper2015recovering} or sonar is blind~\cite{Crestani2015}---or \textit{why} it is wrong---e.g., there was a collision with the environment~\cite{Park2017}. However, the prior work aims to use the diagnoses for autonomous robot recovery from failure or to facilitate debugging by experts. The problem of generating natural language explanations from a fault diagnosis to allow non-experts to help a robot recover remains largely unexplored.

\begin{figure*}[t]
\centering
\includegraphics[width=15cm]{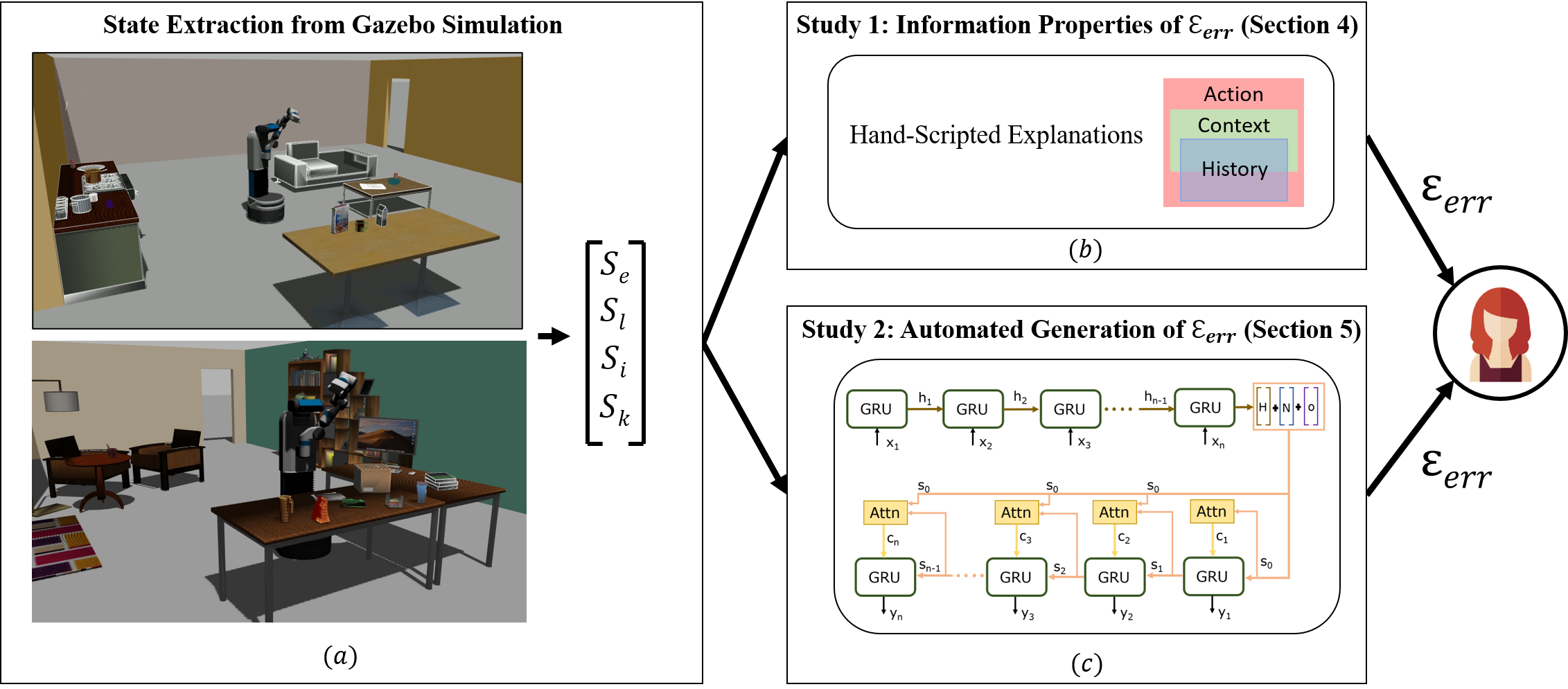}
\vspace{-.2cm}
\caption{The pipeline used to generate $\mathcal{E}_{err}$ explanations for a non-expert user. (a) Data collection in failure simulations and the extraction of the agent's state space. (b) Study of hand-scripted $\mathcal{E}_{err}$ explanations with varying information types (Sec.~\ref{sec:information-properties}). (c)~Autonomously generated $\mathcal{E}_{err}$ explanations using an encoder-decoder model (Sec.~\ref{sec:automated-explanations}).}
\label{fig:sys_arch}
\vspace{-.2cm}
\end{figure*}


In efforts to provide explanations to non-experts on infeasible agent behaviors, prior work has presented a linear temporal logic (LTL) framework to explain actions unsatisfiable by a robot~\cite{raman2012explaining}. The explanations focus on \textit{what} actions are unattainable by the robot, but do not include the underlying reasons for \textit{why} they may be unattainable. Similarly, an algorithm called HIGHLIGHTS uses visual animations to summarize agent capabilities---\textit{what} an agent can achieve---to non-expert users, based on the features dictating an agent's reward function~\cite{amir2018highlights}. In our work, we show that non-experts need to be told why an action is unattainable in order to help an agent recover; explaining what is unattainable is insufficient.


Finally, prior work in XAI has found that natural language explanations can provide ``justification'' and are ``understandable'' by non-experts~\cite{ehsan2019automated}. The study, conducted in the discrete domain of Frogger, used sequence-to-sequence learning to treat explanation generation as a neural translation problem, where an agent's internal states are translated into natural language, with impressive results on non-experts' abilities to comprehend the agent's decisions~\cite{ehsan2018rationalization, ehsan2019automated}. We build on these findings and adapt the sequence-to-sequence learning approach to a continuous robotics domain.

\section{Problem Definition}
\label{sec:problem-definition}

We define the problem of providing \textit{explanations for task failures} by extending the framework introduced by Chakraborti et al.~\cite{chakraborti2020emerging} for producing explanations to goal-directed plans. In the framework, a planning problem $\Pi$ is defined by a transition function $\delta_\Pi : A \times S \rightarrow S \times \mathbb{R}$, where $A$ is the set of actions available to the agent, $S$ is the set of states it can be in, and $\mathbb{R}$ is a cost of making the transition. A planning algorithm $\mathbb{A}$ solves $\Pi$ subject to a desired property $\tau$ to produce a plan or policy $\pi$, i.e. $\mathbb{A} : \Pi \times \tau  \mapsto \pi$. Here, $\tau$ may represent different properties such as soundness, optimality, etc.  The solution to this problem, i.e. the \textit{plan}, $\pi = \langle a_1, a_2, ..., a_n \rangle, a_i \in A$, which transforms the current state $I \in S$ of the agent to its goal $G \in S$, i.e. $\delta_\Pi (\pi, I) = \langle G, \Sigma_{a_i \in \pi} c_i \rangle$. The second term in the output denotes a plan cost $c(\pi)$.

Given the above framework, we define two explanation types.  The first is from~\cite{chakraborti2020emerging}, and the second is contributed by our work:


\begin{flushitemize}
    \item[\textbf{$\mathcal{E}_\pi$}:] This explanation justifies to a human user that solution $\pi$ satisfies property $\tau$ for a given planning problem $\Pi$.  For example, the user may ask ``Why $\pi$ and not $\pi'$?''.  In response to this question, $\mathcal{E}_\pi$ must enable the user to compute $\mathbb{A} : \Pi \times \tau  \mapsto \pi$ and verify that either $\mathbb{A} : \Pi \times \tau  \not \mapsto \pi'$, or that $\mathbb{A} : \Pi \times \tau  \mapsto \pi'$ but $\pi \equiv \pi'$ or $\pi$ is greater than $\pi'$ with respect to some criteria.  $\mathcal{E}_\pi$ applies to the plan solution as a whole and can be elicited at any time.  Approaches addressing $\mathcal{E}_\pi$ are discussed in Sec.~\ref{sec:related-work}.  
    
    \item[\textbf{$\mathcal{E}_{err}$}:] This explanation applies when an unexpected failure state, $f \in\mathcal{F}$, is triggered by a failed action in $\langle a_1, a_2, ..., a_n \rangle$, and halts the execution of $\pi$.  For example, the user may ask ``The robot is at the table, but why did it not pick up my beverage?'' In response to this question, $\mathcal{E}_{err}$ must allow the user to understand the cause of error in order to help the system recover.
\end{flushitemize}

In this work, we develop the second variant of explanations, $\mathcal{E}_{err}$. We assume that both the algorithm $\mathbb{A}$ and the plan $\pi$ are sound, and that the cause of error is triggered by a failure state $f \in \mathcal{F}$ from which an agent cannot recover without user assistance. For example, a situation in which a robot requires human help to discern an occluded object or pickup a tool out of reach. Our objective is to generate $\mathcal{E}_{err}$ such that the user (1) correctly understands the cause of failure, and (2) helps the agent recover from the error by providing a solution.

In the following sections, we present our methods to achieve the above objective. In Sec.~\ref{sec:information-properties}, we introduce a set of \textit{information types}, $\Lambda$, that are characteristics of $\mathcal{E}_{err}$. We then develop scripted explanations satisfying different $\lambda \in \Lambda$, and evaluate them to find a meaningful $\lambda$ that satisfy our objective for non-expert users  (Fig.~\ref{fig:sys_arch}(b)). The results from Sec.~\ref{sec:information-properties} inform our efforts in Sec.~\ref{sec:automated-explanations} to automatically generate $\mathcal{E}_{err}$ without using pre-defined scripts (Fig.~\ref{fig:sys_arch}(c)).


\begin{center}
\begin{table*}[t]
\begin{tabular}{ |p{3.2cm} |c|c|c| p{11.7cm} | }
\hline
 \textbf{Study Condition} & \textbf{$a_t$} & \textbf{$a_{t-1}$} & \textbf{$c_t$} & \textbf{Example Explanation for ``object is occluded'' failure}\\
\hline
 None &  &  &  & N/A \\
 \hline
 Action Based (AB) & $\checkmark$ &  &  & \textit{Robot could not find the object.} \\
 \hline
 Context Based (CB) & $\checkmark$ &  & $\checkmark$ & \textit{Robot could not find the object because the object is hidden from view.} \\
 \hline
 Action Based History (AB-H) & $\checkmark$ & $\checkmark$ &  & \textit{The robot finished scanning objects at its current location but could not find the desired object.} \\
 \hline
 Context Based History (CB-H) & $\checkmark$ & $\checkmark$ & $\checkmark$ & \textit{The robot finished scanning objects at its current location, but could not find the desired object because the desired object is hidden from view.} \\
\hline
\end{tabular}
\caption{The features that can encompass an explanation based on the study conditions. $a_{t}$ represents current action, $a_{t-1}$ represents last successful action, and $c_{t}$ represents captured environmental context.}
\label{tab:Exps}
\vspace{-.7cm}
\end{table*}
\end{center}

\vspace{-.5cm}

\section{Information Types of $\mathcal{E}_{err}$}
\label{sec:information-properties}

In order to generate $\mathcal{E}_{err}$, the first question we have to answer is: \textit{given an error while executing a plan $\pi$ for a particular task, what types of information  should explanation $\mathcal{E}_{err}$ contain}? 

For our application, we desire that $\mathcal{E}_{err}$ is (1) accessible to non-experts, and (2) representative of the fault cause.  Unfortunately, it is not clear from prior literature what information from an agent's plan $\pi$, or its failure, satisfies these requirements.  Ehsan et al.~\cite{ehsan2019automated} propose that explanations for everyday users should take the form of \textit{rationales}, which justify the agent's decision in layperson's terms. However, their rationales are trained from non-expert labels, do not reveal the true decision making process of an agent, and thus would not be able to disambiguate among visually similar robot errors (e.g., failure to grasp object due to kinematic constraints vs. object occlusion vs. a segmentation error).  By contrast, prior work on fault diagnosis~\cite{khalastchi2018fault} has extensively studied how to describe error states, but such work exclusively targets expert users, with resulting explanations referencing specific system components or agent internals (e.g., ``localization mismatch with odometry''~\cite{Crestani2015}).  Thus, our first step is to determine what information $\mathcal{E}_{err}$ should contain to be both accurate and interpretable by non-experts.

In this section, we define a set of information types, $\lambda \in \Lambda$, that we use to generate scripted explanations during a failure. In a user study with non-experts, we determine which $\lambda$ best help users  identify the cause of a failure and suggest solutions to the failure.  Specifically, we conducted a between-subjects user study in which $\Lambda$ consists of four values that are a cross-product of two factors: 2~(history, no history) x 2~(context-based, action-based). In the user study, the four explanation conditions were contrasted against a baseline condition. The five conditions are enumerated below:
\begin{flushitemize}
    \item \textbf{Baseline (None)}: Participants receive no explanation on the cause of error. This is the current standard in deployed robotic systems, e.g.,~\cite{sauppe2015social}.
    \item \textbf{Action-Based (AB)}: Participants receive {$\mathcal{E}_{err}$} containing only the currently failed action $a_{t}$ as the cause of error. 
    \item \textbf{Context-Based (CB)}: Participants receive {$\mathcal{E}_{err}$} containing both $a_{t}$ and context, $c_{t}$, retrieved from the environment as the cause of error.
     \item \textbf{Action-Based-History (AB-H)}: Participants receive {$\mathcal{E}_{err}$} containing the previous action $a_{t-1}$ and $a_{t}$ as the cause of error.
    \item \textbf{Context-Based-History (CB-H)}: Participants receive {$\mathcal{E}_{err}$} containing $a_{t-1}$, $a_{t}$, and $c_{t}$ as the cause of error.
\end{flushitemize}
Table~\ref{tab:Exps} summarizes the study conditions and provides example explanations for each condition. In the following sections, we discuss our experimental setup, the study procedure, and our results.



\subsection{Goal, Action Space, \& State Space}
\label{sec:properties-state-space}

We conduct our experiment in Gazebo simulations of a Fetch robot~\cite{wise2016fetch}. The Fetch robot is a mobile manipulator with a differential drive base, a 7DoF arm, a parallel-jaw gripper, a pan-tilt head, and an adjustable torso. For sensing, the base includes a laser scanner and the head contains an RGB-D camera. The robot is simulated in a kitchen setting performing a pick-and-place task (as seen in Fig.~\ref{fig:sys_arch}a). The robot's task is to move a task-specified object (e.g., milk carton) from the dining table to the kitchen counter.


Similar to prior work in robotics~\cite{banerjee2020taking}, we define the robot's action space as the set $A = \{move, segment, detect, findgrasp, grasp, lift,\\place\}$, where $move$ navigates the robot to a specified location, $segment$ is used to identify which pixels in its sensory space correspond to objects, $detect$ performs object detection to obtain a label for a given object, $findgrasp$ executes grasp sampling to identify possible grasp poses for the gripper, $grasp$ moves the robot arm into a grasp pose and closes the gripper, $lift$ raises the arm, and $place$ places a held object at a specified location.

The robot's state at each time step $t$ is defined as $s_{t} \in S$, where $S = S_{e} \cup S_{l} \cup S_{i} \cup S_{k}$. Here, $S_{e} = S_{o} \cup S_{p}$ denotes the set of names for all entities in the environment, where $S_{o}$ consists of \textit{\{milk, coke can, ice cream, bottle, cup\}}, and $S_{p}$ consists of: \textit{\{dining table, kitchen counter\}}. $s_{l}(t) \in S_{l}$ is a vector of $\langle x , y, z \rangle$ locations for each entity $s_{e} \in S_{e}$ at a given time step $t$. $s_{i}(t) \in S_{i}$ is defined by three tuples $\langle x_{avel},y_{avel},z_{avel}\rangle$, $\langle x_{lvel},y_{lvel},z_{lvel}\rangle$ ,$\langle x_{pos},y_{pos},z_{pos} \rangle$ that describe the angular velocity, linear velocity and position of the robot at $t$. Finally, $S_{k} = \{k_{grasp},k_{findgrasp},k_{move},k_{pick},k_{detect},k_{seg}\}$ where $s_{k}(t) \in S_{k}$ describes the status of each $a \in A$ at $t$, and whether each action is: active ($0$), completed ($1$) or errored (-$1$). Therefore, at all time steps, the number of elements in $s_{k}(t)$ is equal to the number of actions in $A$.


\vspace{-.2cm}
\subsection{Failure Taxonomy}
\label{sec:simulating-failures}

The agent's initial state is defined as $s_{0} = \{\langle 0,0,0\rangle,\langle 0,0,0\rangle, \langle 0,0,0\rangle,\\\{null\}\}$, where the position tuple and the velocity tuples are set to zero, and the action states $s_{k}(0)$ are not defined. If there are no errors, the agent's final state is defined as $s_{T} = \{\langle x_{T},y_{T}, z_{T}\rangle, \langle 0,0,0\rangle,\\\langle 0,0,0\rangle, \{1,1, ..., 1\}\}$, where the position tuple is set to the goal location, the velocity tuples are zero, and each action state in $s_{k}(T)$ is 1. In this context, plan $\pi$ is the set of actions $\langle a_1, a_2, ..., a_n \rangle \in A$ that transform the agent's initial state $s_{0}$ to its final state $s_{T}$. We then define a failure $f$ in plan $\pi$ as the event when any action state in $s_{k}$ has a value -1.

\begin{figure}[t]
\centering
\includegraphics[width=\linewidth]{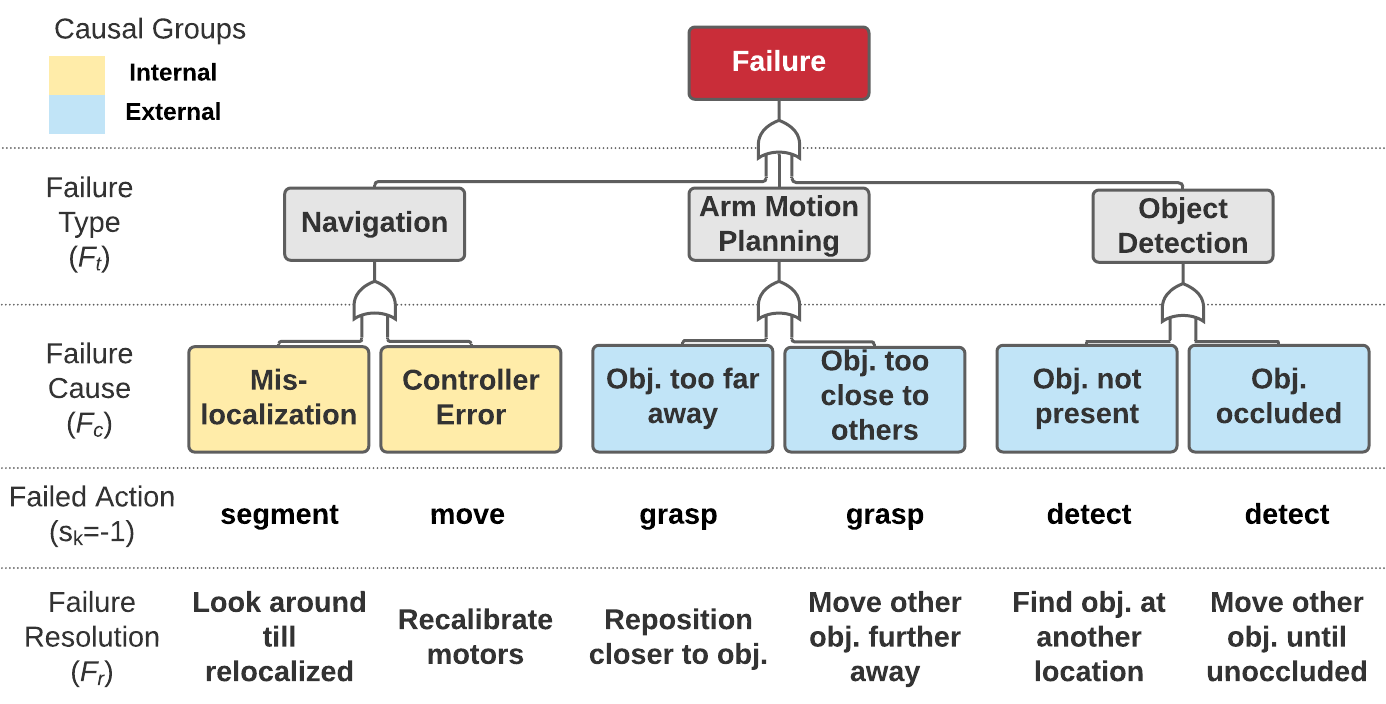}
\caption{Fault tree analysis of failures in this work. We also show the failed action used to detect the failure, and a shorthand label of the solution to fix the failure.}
\label{fig:fault-taxonomy}
\vspace{-.5cm}
\end{figure}

Following the example of prior work~\cite{Park2017}, we study our work in the context of a representative sample of failures in robot behavior.  We classify these failures using fault-tree analysis (Fig.~\ref{fig:fault-taxonomy})\footnote{Our fault-tree analysis identifies errors and solutions relevant to our domain, and we leave generating explanations for unknown errors for future work. Visualization of each failure is available at: \url{https://youtu.be/jYn3FaqG65E.}}. Failed robot behaviours characterize coarse failure types ($F_t$), e.g., a failure in ``object detection''. Each failure type can have multiple failure causes ($F_c$), e.g., ``object not present'' or ``object occluded'' are possible causes for an ``object detection'' failure. In our system, failures are detected by an errored action. For example, $s_{k_{detect}} = -1$ can indicate that either the ``object is occluded'' or the ``object is not present''. Crucially, however, each failure cause has an associated resolution action, not in the robot's action space, but which can be selected by humans to rectify the cause of failure.


The fault-tree analysis also groups failure causes $F_c$ into causal groups, which we define as \textit{Internal} and \textit{External}.  These categories roughly correspond to system and environment failures, respectively, in the prior work~\cite{Park2017}. Internal failures are not apparent through visual cues in the environment and are often the result of failures of hardware or software modules. By contrast, external failures are often caused by unexpected conditions in the environment and are therefore visually apparent in the environment. In Section~\ref{sec:properties-results}, we investigate the effect of different information types on users when the error stems from the different causal groups.

\begin{figure*}[!t]
\centering
\begin{subfigure}[b]{0.244\textwidth}
  \centering
  \includegraphics[width=\textwidth]{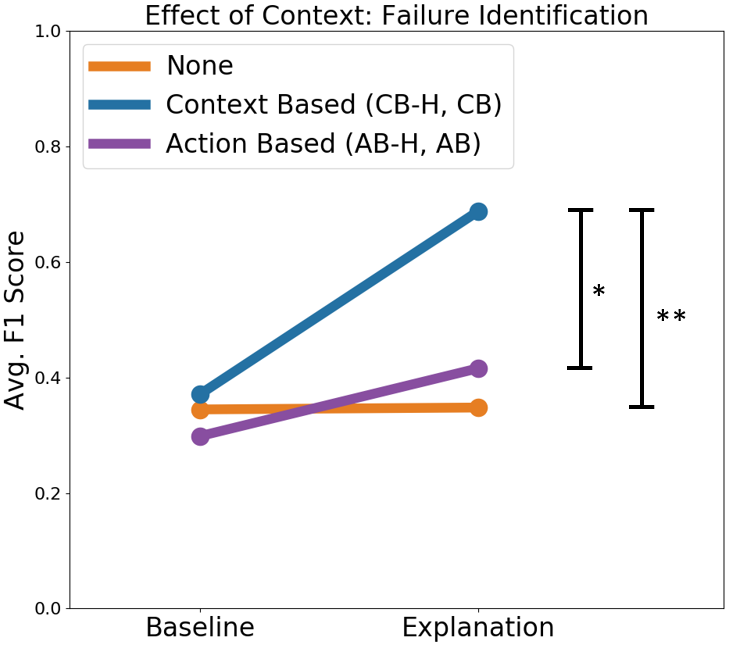}
  \caption{}
  \label{fig:analysis-ac-fid}
\end{subfigure}
\hfill
\begin{subfigure}[b]{0.245\textwidth}
  \centering
  \includegraphics[width=\textwidth]{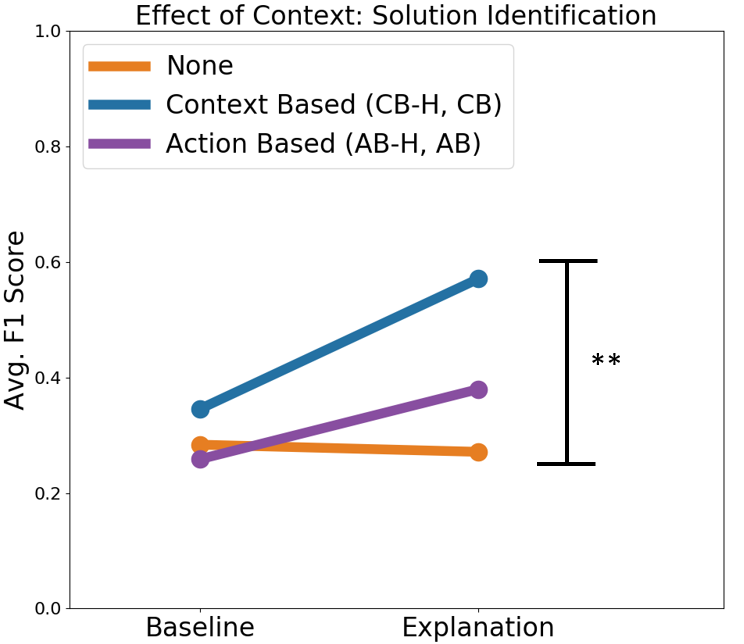}
  \caption{}
  \label{fig:analysis-ac-sid}
\end{subfigure}
\hfill
\begin{subfigure}[b]{0.245\textwidth}
  \centering
  \includegraphics[width=\textwidth]{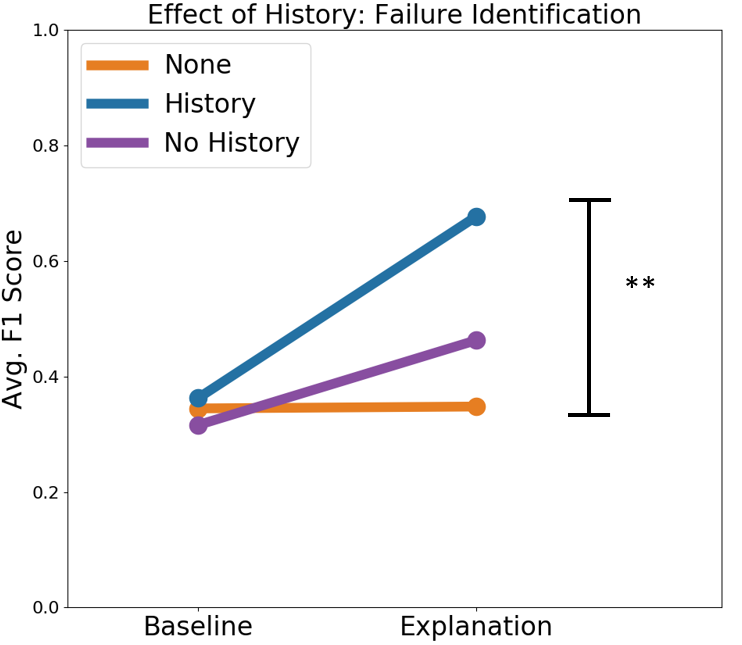}
  \caption{}
  \label{fig:analysis-hnh-fid}
\end{subfigure}
\hfill
\begin{subfigure}[b]{0.238\textwidth}
  \centering
  \includegraphics[width=\textwidth]{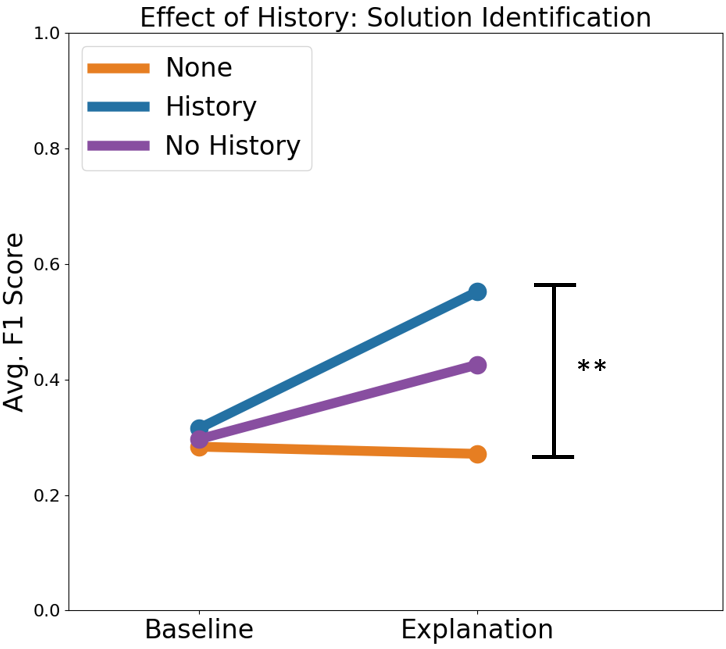}
  \caption{}
  \label{fig:analysis-hnh-sid}
\end{subfigure}
\vspace{-.3cm}
\caption{Average F1 score across explanation conditions grouped by Context Based vs. Action Based (a-b) and History vs. No History (c-d). In Fig.~\ref{fig:ANALYSIS}, \ref{fig:INT-EXT}, and \ref{fig:ANALYSIS_2}, statistical significance is reported as: *p < 0.05, **p < 0.01, ***p < 0.001}~\label{fig:ANALYSIS}
\vspace{-.3cm}
\end{figure*}

\begin{figure*}[!t]
\centering
\begin{subfigure}[b]{0.244\textwidth}
  \centering
  \includegraphics[width=\textwidth]{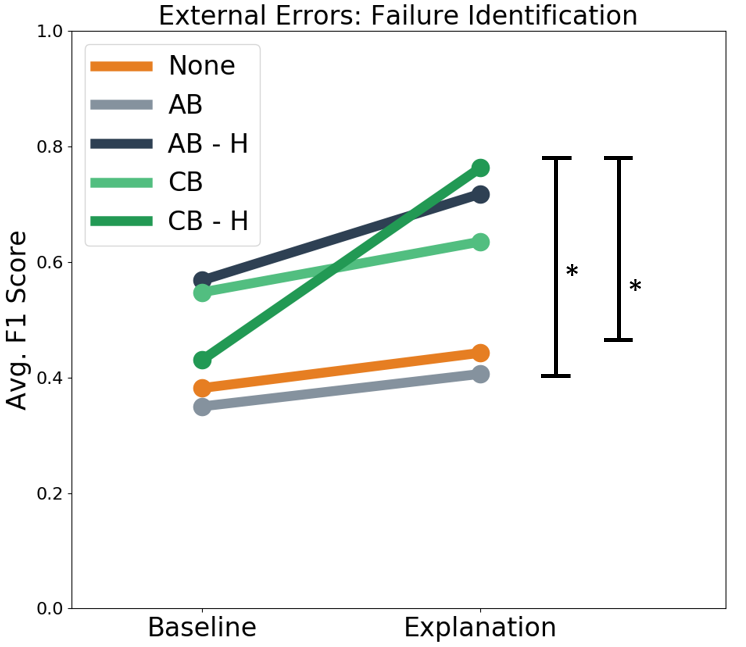}
  \caption[]%
  {{}}
\end{subfigure}\quad
\begin{subfigure}[b]{0.245\textwidth}
  \centering
  \includegraphics[width=\textwidth]{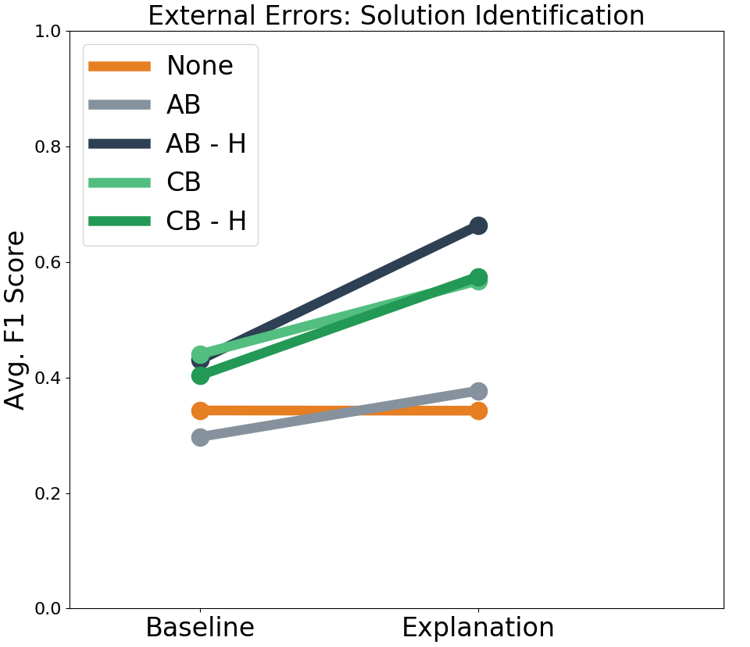}
    \caption[]%
     {{}}
     \end{subfigure}
\begin{subfigure}[b]{0.245\textwidth}
  \centering
  \includegraphics[width=\textwidth]{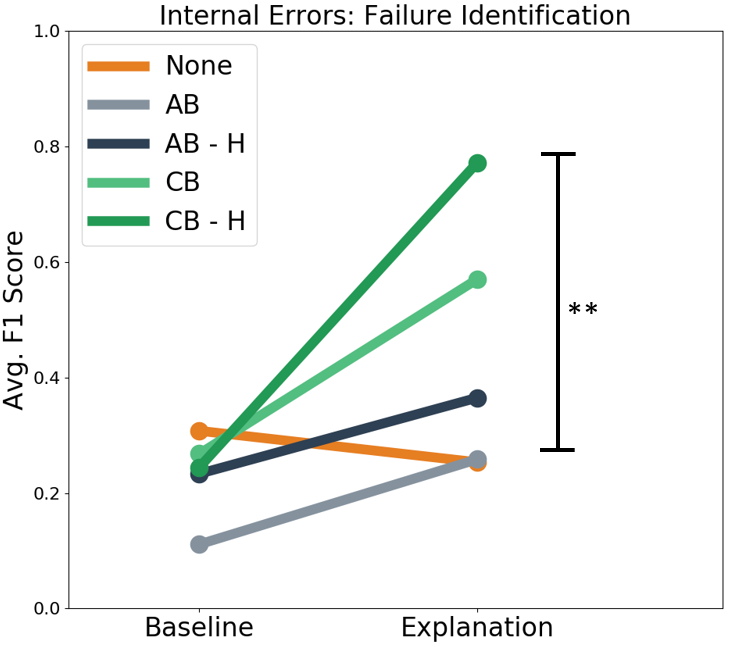}
    \caption[]%
     {{}}
     \end{subfigure}
\begin{subfigure}[b]{0.238\textwidth}
  \centering
  \includegraphics[width=\textwidth]{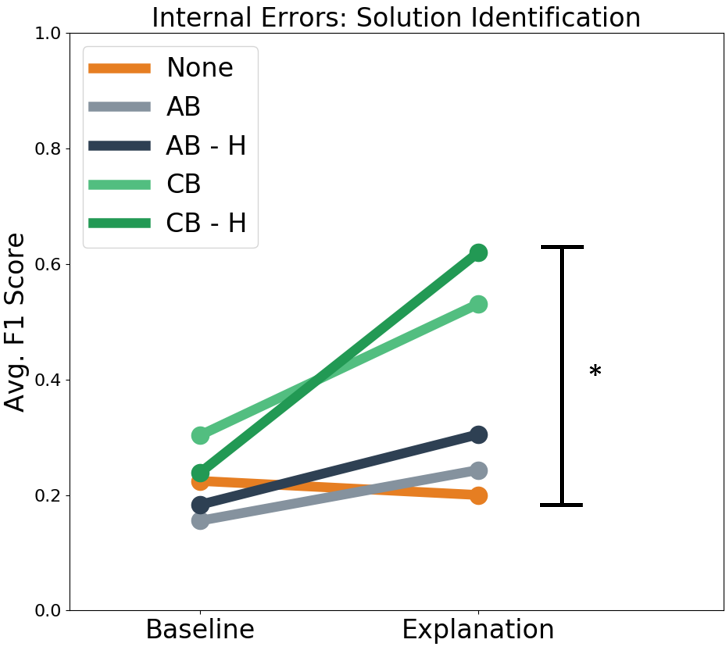}
    \caption[]%
     {{}}
     \end{subfigure}
\vspace{-.6cm}
\caption{Average F1 score across all conditions grouped by Internal versus External errors.}
\label{fig:INT-EXT}
\vspace{-.3cm}
\end{figure*}

\subsection{Study Design}
\label{sec:properties-study-design}

Our objective is to evaluate the different information types of error explanations across a variety of failures. We simulated $|\mathcal{F}| = |S_o| \times |F_c| = 30$ failures to capture all possible object $\times$ failure cause combinations. In our domain, each failure $f \in \mathcal{F}$ has a single cause in $F_c$ and therefore a single resolution method $F_{r}$. The study consisted of the following three stages.

\noindent \paragraph{Familiarization:} Participants in all conditions were first shown three videos of the Fetch robot successfully executing the task with randomly selected objects from $S_o$ using a plan $\pi$.  This served to accustom participants to the robot, its abilities, and its actions. 

\noindent \paragraph{Baseline:} All participants were then shown six randomly sampled failure simulations from $\mathcal{F}$, one for every failure cause $F_c$. To visualize the failure, participants were shown animated snapshots (GIFs) of actions leading up to a failure, and three perspective shots of the robot in the final environment state\footnote{Humans subject study is available here: \url{https://robotasks00.web.app/.}}. Participants were provided \textbf{no explanations} and asked to identify the cause of the failure and suggest a solution. Participant responses established the participants' baseline understanding of the robot and the domain, allowing us to measure improvement in understanding.

\noindent \paragraph{Explanation:} Finally, participants were exposed to twelve additional randomly sampled failures from $\mathcal{F}$ (different from \textit{Baseline}), two for every failure cause $F_c$. Depending on the assigned study condition, a participant was either provided a hand-scripted explanation matching the information type of the assigned study condition, or the participant was provided no explanation if in the \textit{None} condition. As before, the participant was required to identify the cause of failure and suggest a solution.
For each simulation, after identifying a failure and solution, participants received their accuracy score. This was the only feedback given to all participants.

\subsection{Measures \& Hypotheses}
\label{sec:properties-metrics-hypotheses}



We evaluate participant performance using F1 score. In particular, we evaluate the difference between participant \textit{Baseline} F1 score and their \textit{Explanation} F1 score. The difference in F1 score is evaluated for the following measures:
\begin{flushitemize}
    \item \textbf{Failure Identification (\textit{FId})}: measures a participants' ability to correctly identify the cause of each failure.
    \item \textbf{Solution Identification (\textit{SId})}: measures a participants' ability to correctly identify the solution to each failure.
\end{flushitemize}

Our data analysis then aims to answer the following questions with respect to the measures: 
\begin{flushitemize}
    \item \textbf{Q1}: Do action-based (AB) or context-based (CB) explanations lead to the greatest improvement in user failure identification (\textit{FId}) and solution identification (\textit{SId})?
    \item \textbf{Q2}: Does the inclusion of history within an explanation improve users' failure identification (\textit{FId}) and solution identification (\textit{SId})?
    \item \textbf{Q3}: How do users' failure identification (\textit{FId}) and solution identification (\textit{SId}) compare for Internal vs External robot errors?
\end{flushitemize}

\subsection{User Evaluation of Scripted Explanations}
\label{sec:properties-results}

\textbf{Participants}.
We recruited 80 individuals from Amazon's Mechanical Turk. Since our target audience is non-experts, we filtered out 10 participants for achieving 100\% accuracy in the \textit{Baseline} stage, under the assumption that they were not novices. The remaining 70 participants included 51 males and 19 females, all whom were 18 years or older (M = 35.2 , SD = 9.4). Due to the exclusion criteria, each study condition had 13-15 participants. The task took on average 20 - 40 minutes and participants were compensated \$3.50.

\noindent\textbf{Data Analysis}.
The data on the \textit{FId} and \textit{SId} metrics are analyzed with a two-way ANOVA for \textbf{Q1} and \textbf{Q2} and a one-way ANOVA for \textbf{Q3}, followed by a Tukey HSD post-hoc test for each.

\smallskip
\noindent\textbf{Fig.~\ref{fig:analysis-ac-fid} and Fig.~\ref{fig:analysis-ac-sid} answer Q1} by showing the benefit of including environmental context (CB, CB-H conditions) in failure identification (\textit{FId}) and solution identification (\textit{SId}). In both figures, we see that explanations with context have the highest improvement in \textit{FId} and \textit{SId} scores. Specifically, the presence of context had a significant effect on \textit{FId} (F(2,67)= 6.95, p=0.0018), with a significant \textit{FId} improvement for Context-based explanations over both None (t(67)=3.729, p=0.0012) and Action-based (t(67)=2.923,p=0.014) explanations. Similarly, the presence of context had a trending effect on \textit{SId} (F(2,67)=2.92, p=0.06), with a significant improvement in \textit{SId} for Context-based explanations vs. None (t(67)=3.12, p=0.007). This indicates that the inclusion of environmental context in the CB explanation conditions (CB, CB-H) helped participants better understand the underlying causes of the failures thereby allowing them to better assist the robot.

\smallskip
\noindent\textbf{Fig.~\ref{fig:analysis-hnh-fid} and Fig.~\ref{fig:analysis-hnh-sid} answer Q2} by showing the benefit of including history (AB-H, CB-H conditions), on \textit{FId} and \textit{SId}. In both figures, history-based explanations have the highest improvement in \textit{FId} and \textit{SId} scores. Similar to the effects of including context, including history had a significant improvement on \textit{FId} (F(2,67)= 3.36, p=0.04), with History vs. None  as significant (t(67)=3.447, p=0.003). Although including history did not have a significant effect on \textit{SId} overall (F(2,67)= 1.38, p=0.25), we observe a significant difference in improvement between History-based explanations vs. None (t(67)=3.1857, p=0.006). This supports the idea that knowledge of the most recently completed action (AB-H, CB-H conditions) can help users gauge what a robot was \textit{able} to successfully accomplish, thereby helping users better pinpoint the exact cause of failure and provide correct suggestions for recovery.

\smallskip
Our analysis so far investigates the independent effects of including context and history on explanation utility. The results suggest that context-based explanations incorporating history, i.e. CB-H explanations, are the best suited to non-experts. We next consider each explanation type individually and their efficacy for non-experts based on the causal group of the originating fault. 

\smallskip
\noindent\textbf{Fig.~\ref{fig:INT-EXT} answers Q3} by showing the different effects of the explanation types for failures stemming from the different causal groups---\textit{Internal} and \textit{External} failures. Explanations have a significant effect on the improvement in \textit{FId} for \textit{External} errors (F(4,62)= 3.53, p=0.01), with CB-H showing the most pronounced improvement, specifically vs. AB (t(62)=-3.216, p=0.017) and vs. None(t(62)=-3.046, p=0.027). Additionally, we see a significant effect of explanations in improving \textit{FId} for \textit{Internal} errors ((F(4,62)= 4.39, p=0.003), with a significant difference in CB-H vs. None (t(62)=-3.955, p=0.0018). With respect to improvement in \textit{SId} for \textit{External} errors, we see a trending effect of explanations (F(4,62)=2.16, p=0.083) with a trending difference between AB-H vs. None (t(62)=2.648, p=0.073). For \textit{Internal} errors, we notice a trending effect of explanations (F(4,62)=2.37, p=0.061), but with a significant difference between CB-H and None (t(62)=-2.86, p=0.044). Overall, we find that CB-H explanations are valuable to participants for both error types, but especially for the Internal case when failure causes are not discernible through the environment.

\section{Automated Generation of $\mathcal{E}_{err}$}
\label{sec:automated-explanations}
In Sec.~\ref{sec:information-properties} we discovered CB-H explanations to be most effective. In this section, we introduce an automated explanation generation system that can generate the CB-H \footnote{The system is also able to generate AB, AB-H, and CB explanations; but we focus on CB-H due to its highest \textit{FId} and \textit{SId} scores in Sec.~\ref{sec:properties-results}.} explanations word by word, without a template. 

\subsection{Encoder-Decoder Model Overview}
\label{sec:automated-model-overview}

We adapt a popular encoder-decoder network \cite{bahdanau2015neural,bastings2018annotated} utilized by \cite{ehsan2019automated} to train a model to generate CB-H explanations from an agent's state. The model's features, $U$, are derived from the state space, $S$ (see Sec.~\ref{sec:feature-set}), and are comprised of environment features $X$, continuous features, $N$, and a desired object of interest, $o$. The encoder receives the environment features as input and produces an embedding of the environment context in its hidden state, $h_n$. The embedding is then appended to the continuous features and the object of interest, and the concatenated features are given to the decoder as input. The decoder generates a sequence of target words, $Y= \{y_{1},y_{2}...y_{m}\}$, where $y_{i}$ is a single word, and $Y$ is the CB-H explanation. The model architecture is shown in  Fig.~\ref{fig:sys_arch}(c).


The encoder and decoder are comprised of Gated Recurrent Units ($GRU$) \cite{cho2014learning}. Given a sequence of environment features, $X = \{x_{1},x_{2}...x_{n}\}$, the encoder generates the context embedding at sequence step $i$ by, $h_{i} = GRU(x_{i},h_{i-1})$, where $h_{i-1}$ is the previous step's context embedding. The decoder uses the final context embedding, $h_n$, concatenated to the continuous features, $N$, and the object of interest, $o$, as its initial input, $s_0$. The decoder also generates and uses a weighted attention vector, $c_i$, for step i (initialized with $c_0 = 0$). At each step, $c_i$ attends over the features in $s_0$ and $s_{i-1}$, the decoder's input at the previous step. The decoder then updates its state according to the function $s_i = GRU(s_i, y_{i-1}, c_i)$, where $y_{i-1}$ is the previous predicted word, and a word, $y_i$, is predicted from the maximum softmax probability over $s_i$. A complete explanation is generated when the decoder predicts the `END' token.


\begin{figure}[]
\centering
  \includegraphics[width=1\columnwidth]{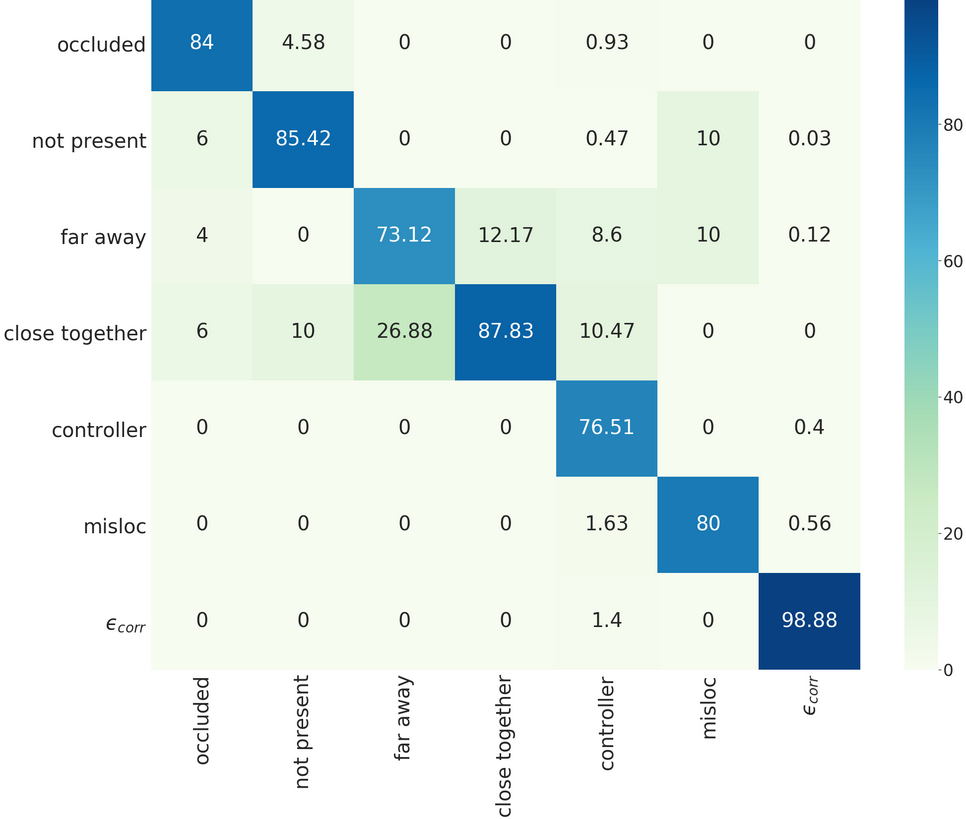}
  \caption{Confusion matrix analysis of our model's performance where the first six columns represent $\mathcal{E}_{err}$ explanations and the last column represents  $\mathcal{E}_{corr}$ rationalizations. The x-axis represents the true labels, and the y-axis represents the predicted labels.}~\label{fig:CM}
\vspace{-.8cm}
\end{figure}

\subsection{Feature Set}
\label{sec:feature-set}

Recall from Sec.~\ref{sec:properties-state-space} that the agent's state space is defined as $S = S_{e} \cup S_{l} \cup S_{i} \cup S_{k}$. We derive the features, $U = X \cup N \cup o$ for the encoder-decoder model from $S$. The object of interest, $o \in S_e$, is specified as part of the task and represented by its word embedding in $U$. The environment, $X$, is comprised of the word embeddings of the names of the objects, $Obj_G$, located in the robot's area of interest, such that $\forall o' \in Obj_G,\; o' \in S_e$. The remaining continuous features, $N = \{Rel_{a-Goal},Rel_{a-o}, v_{ang}, v_{lin},  S_{k},Rel_{o-Obj_{G}},o_{p}\}$, characterize the robot and the target object in the environment. $Rel_{a-Goal}$ is the distance of the robot from its goal location, $Rel_{a-o}$ is the distance of the robot from the target object, $v_{ang}$ and $v_{lin}$ are the angular and linear velocities of the robot base, and $S_k$ are the action statuses as defined in Sec.~\ref{sec:properties-state-space}. Additionally, $Rel_{o-Obj_G}$ is the distance between the desired object $o$ and the objects in $Obj_G$, and $o_p$ is a boolean that evaluates to true if $o \in Obj_G$. Note that not all the features in $N$ contain valid values at all times. If a feature value is invalid, the feature is masked before it is concatenated into $U$.


\vspace{-.2cm}
\subsection{Data Collection}
\label{sec:data-collection}

To further evaluate the generalizability of our method across environments, we expand our data to include simulations from an office environment (Fig.~\ref{fig:sys_arch}(a)), in addition to the kitchen environment introduced in Sec.~\ref{sec:information-properties}. The office environment contains different objects and locations, i.e. entities $S_e$, but the robot's pick-and-place task remains the same. Entities in the office environment have a one-to-one correspondence to the entities in the kitchen environment. 



Our dataset $D$ consists of 72 simulations (60 and 12 from the kitchen and office environments, respectively).
Each timestep in $D$ is defined by $u_{t}$, where $u_{t} \in U$ represents the input features to our encoder-decoder model at timestep $t$. Each simulation begins with $n$ active or successful action timesteps, denoted by $s_{k}(t) = 0$ or $s_{k}(t) = 1$, and ends with $m$ error timesteps, denoted by $s_{k}(t) = -1$. In our work, $n$ ranges from 15 to 20, and $m=10$. The $m$ error timesteps simulate a robot repeatedly attempting to autonomously remedy a failure upon encountering it, reflecting a real world solution to errors where robots try to repeat actions that fail~\cite{banerjee2020taking}.

Given our dataset, we annotate error timesteps with a CB-H explanation, $\mathcal{E}_{err}$, and annotate successful or active timesteps with a natural language rationalization of the state, $\mathcal{E}_{corr}$, as in~\cite{ehsan2018rationalization}. In our work, examples of such rationales include, ``robot moving to dining table'' and ``robot segmented objects in the scene''. Additionally,  $\mathcal{E}_{corr}$ explanations were only used for model training, and were not a focus of the human subjects study in Sec.~\ref{sec:model-user-eval}. The total size of $D$ is 2100 timesteps where there are 1380 successful or active timesteps, and 720 error timesteps.

\vspace{-.1cm}
\subsection{Model Training \& Evaluation}
\label{sec:training-testing}

Our encoder-decoder model is trained with the 60 kitchen simulations using a two-step grouped leave one out cross validation (LOOCV) with 10 folds, where the grouped LOOCV leaves out an entire simulation for each failure cause, $F_c$. The first grouped LOOCV creates a split between the training set, $d_{tr}$, and test set, $d_{te}$, while the second LOOCV creates the validation set, $d_v$. As a result, in each fold, $d_{tr}$ includes 48 simulations with 480 error explanations, while $d_{v}$ and $d_{te}$ include 6 simulations, each with 60 error explanations. To evaluate each fold, we utilize an evaluation set, $d_{eval}$, which includes the 12 office simulations with 120 error explanations.

\noindent\textbf{Training.} Our models trained for an average of 180 epochs, depending on the validation loss. We train with a batch size of 20. Our GRU cells in the encoder have a hidden state size of 20 and the GRU cells in the decoder have a hidden state size of 49. We train our model using a Cross Entropy loss optimized via Adam with a learning rate of 0.0001.

\noindent\textbf{Evaluation}. Fig.~\ref{fig:CM} shows the average performance of the model on $d_{eval}$ across the 10 folds of cross-validation. The confusion matrix includes accuracy on explanations, $\mathcal{E}_{err}$, for the six failure causes as well as accuracy on the non-error rationalizations, $\mathcal{E}_{corr}$. An explanation or rationalization is marked correct only if it identically matches its target phrase.


On average, our model can generalize explanations across the six failure causes with 81.81\% accuracy. For each failure scenario, the model has a larger true positive rate than false positive rate or false negative rate. We observe that the model is accurate in determining ``arm motion planning'' failures but struggles to differentiate between its causes: ``object too far away'' and ``object too close to others'' (Fig.~\ref{fig:fault-taxonomy}). We also notice that explanations of the ``navigation'' and ``object detection'' failure types sometimes indicate causes they are not associated: e.g., ``controller error'' wrongly predicted as ``object too far away'' or ``object too close to others'', or ``object occluded'' wrongly predicted as ``object too far away'' or ``object too close together''. We suspect that the challenges stem from our model's continuous feature space, making certain features harder to distinguish and that with additional training data, the generalizability of our model can be improved.



\noindent\textbf{Model Selection.} Of the 10 models trained with LOOCV, we selected the best model based on its performance on $d_{eval}$. The best model was deployed in a user evaluation described below.


\begin{figure}[t]
\centering
\begin{subfigure}[b]{0.49\linewidth}
  \centering
  \includegraphics[width=\linewidth]{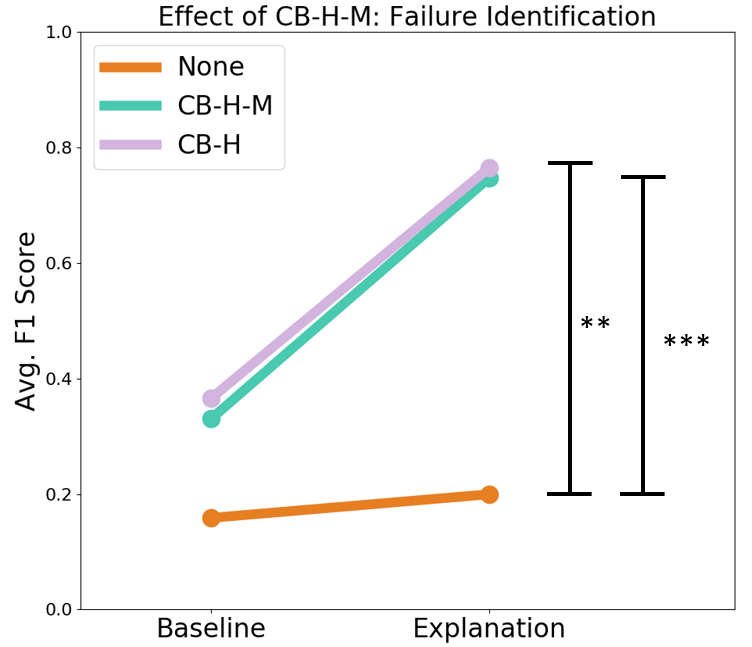}
  \caption{}
\end{subfigure}
\hfill
\begin{subfigure}[b]{0.49\linewidth}
  \centering
  \includegraphics[width=\linewidth]{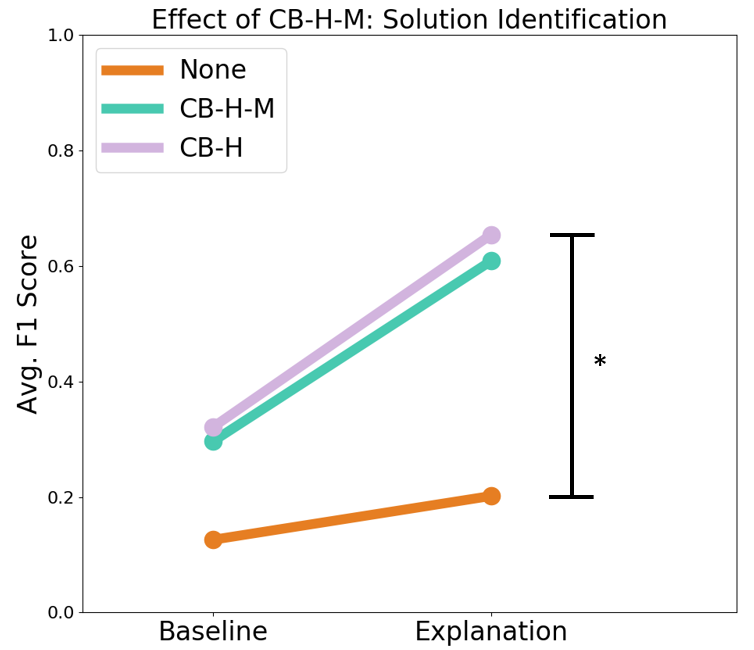}
  \caption{}
\end{subfigure}
\vspace{-.3cm}
\caption{Average F1 scores between participants who received model generated explanations (CB-H-M), scripted explanations (CB-H), and no explanations (None).}
\label{fig:ANALYSIS_2}
\vspace{-.5cm}
\end{figure}

\subsection{User Evaluation of Model Explanations}
\label{sec:model-user-eval}

We conducted a user evaluation similar to the one described in Sec.~\ref{sec:information-properties}. The study was a three condition between subjects study, where participants were either provided with no explanations of errors (\textbf{None}), context-based-with-history scripted explanations (\textbf{CB-H}), or context-based-with-history model-generated explanations (\textbf{CB-H-M}). During the study, participants were shown kitchen simulations in the \textit{Baseline} portion of the study, and evaluated on the 12 office simulations in the \textit{Explanation} portion of the study.

\smallskip
\noindent\textbf{Hypotheses.} We wished to evaluate whether (1) the model generated explanations improved participants' failure and solution identification compared to the None condition, and (2) the model generated explanations performed on par with the hand-scripted explanations in improving participants' performance.

\smallskip
\noindent\textbf{Participants.} We recruited 45 individuals from Amazon's Mechanical Turk.  After applying the exclusion criteria as before, the remaining 41 participants included 25 males and 16 females, all whom were 18 years or older (M = 39 , SD = 11.3). Due to the exclusion criteria, each study condition had 12-15 participants. The task took roughly 20 - 40 minutes and participants were compensated \$3.50.

\smallskip
\noindent\textbf{Data Analysis.} 
The data on \textit{FId} and \textit{SId} metrics are analyzed with a one-way ANOVA followed by a Tukey HSD post-hoc test.

\smallskip
\noindent \textbf{Fig.~\ref{fig:ANALYSIS_2} answers our hypotheses} by showing that CB-H-M explanations are just as effective as CB-H scripted explanations. We observe
a significant effect of explanations on \textit{FId} (F(2,38)=10.52, p=0.0002) and \textit{SId}((F(2,37)=3.94, p=0.027). With respect to \textit{FId} we observe a significant improvement in participant accuracy between CB-H-M and None (t(38)=-4.158,p=0.00049), and no significant difference between CB-H and CB-H-M (t(38)=-0.208, p=0.97). With respect to \textit{SId} we observe a trending difference in improvement between CB-H-M vs. None (t(37)=2.354, p=0.060), a significant difference between CB-H vs, None (t(37)=2.561, p=0.038) and no significant difference between CB-H and CB-H-M (t(37)=0.215, p=0.974). Thus, we conclude that given CB-H-M explanations, participants perform just as well in helping the robot as when given CB-H explanations.

\section{Conclusion \& Discussion}
In this work, we investigate what types of information within an explanation help \textit{non-experts} identify robot failures and help assist in recovery. We introduce a new type of explanation, $\mathcal{E}_{err}$, which has not been previously addressed in the XAIP community, and which describes the cause of unexpected failures amidst plan execution. Our results indicate that for explanations to improve failure and solution identification, they should encompass both environmental context and history of past successful actions. Furthermore, in our first user evaluation we showcase the importance that context-based-history explanations serve in the cases of \textit{Internal} errors, which are not visually observable through environmental changes. Additionally, we investigate a method to autonomously generate such explanations, and verify that they are as effective as its scripted counterpart and  generalizable across environments.

Our work brings XAI techniques into the domain of fault recovery and aims to aid non-expert users (1) understand unexpected failures of a complex robot system and (2) provide recovery solutions in such an event. Although our work includes important contributions, there are limitations that should be addressed by future work. First, while the context-based-history explanations are useful for assisting in failure recovery, they are not guaranteed to be useful to \textit{all} non-experts. Therefore future work can explore tailoring explanations to individual users, perhaps with the reinforcement learning techniques used in recommender systems~\cite{wang2018reinforcement}. Second, our work has characterized the utility of context and history in providing meaningful $\mathcal{E}_{err}$, but we have assumed that explanations can be arbitrarily long. Future work should investigate additional factors that characterize a good $\mathcal{E}_{err}$, and the tradeoffs of providing more information vs. remaining concise. Finally, while the current encoder-decoder model can generalize over varying failure scenarios, there is still room to improve its generalizability to additional situations. Future work can investigate a wider range of simulation domains, tasks, and failures.


\label{sec:conclusion}



\section{Acknowledgments}
This material is based upon work supported by the NSF Graduate
Research Fellowship under Grant No. DGE-1650044.

\bibliographystyle{ACM-Reference-Format}
\bibliography{references}
\end{document}